\documentclass[10pt,letterpaper]{article}
\usepackage[top=0.85in,left=2.75in,footskip=0.75in]{geometry}

\usepackage{amsmath,amssymb}
\usepackage{bm,url,booktabs,algorithm,algorithmic,amsmath,amsfonts,amsthm,mathtools}

\usepackage{changepage}

\usepackage{textcomp,marvosym}

\usepackage{cite}

\usepackage{nameref,hyperref}

\usepackage[right]{lineno}

\usepackage[nopatch=eqnum]{microtype}
\DisableLigatures[f]{encoding = *, family = * }

\usepackage[table]{xcolor}

\usepackage{array}

\newcolumntype{+}{!{\vrule width 2pt}}

\newlength\savedwidth

\raggedright
\usepackage[aboveskip=1pt,labelfont=bf,labelsep=period,justification=raggedright,singlelinecheck=off]{caption}

\makeatletter
\renewcommand{\@biblabel}[1]{\quad#1.}
\theoremstyle{plain}

\newtheorem*{thm*}{Theorem}
\theoremstyle{definition}
\newtheorem{dfn}{Definition}
\makeatletter
\newcommand{\opnorm}{\@ifstar\@opnorms\@opnorm}
\newcommand{\@opnorms}[1]{%
  \left|\mkern-1.5mu\left|\mkern-1.5mu\left|
   #1
  \right|\mkern-1.5mu\right|\mkern-1.5mu\right|
}
\newcommand{\@opnorm}[2][]{
  \mathopen{#1|\mkern-1.5mu#1|\mkern-1.5mu#1|}
  #2
  \mathclose{#1|\mkern-1.5mu#1|\mkern-1.5mu#1|}
}
\makeatother

\makeatother

\usepackage{lastpage,fancyhdr,graphicx}
\usepackage{epstopdf}
\fancyheadoffset[L]{2.25in}
\fancyfootoffset[L]{2.25in}
\begin{document}
\vspace*{0.2in}

\begin{flushleft}
{\Large
\textbf\newline{DC Algorithm for Estimation of Sparse Gaussian Graphical Models} 
}
\newline
\\
Tomokaze Shiratori\textsuperscript{1},
Yuichi Takano\textsuperscript{2}
\\
\bigskip
\textbf{1} Graduate School of Science and Technology, University of Tsukuba, Tsukuba, Ibaraki, Japan
\\
\textbf{2} Faculty of Engineering, Information and Systems, University of Tsukuba, Tsukuba, Ibaraki, Japan
\\
\bigskip

%
%


\end{flushleft}

\section*{Abstract}
Sparse estimation for Gaussian graphical models is a crucial technique for making the relationships among numerous observed variables more interpretable and quantifiable. Various methods have been proposed, including graphical lasso, which utilizes the $\ell_1$ norm as a regularization term, as well as methods employing non-convex regularization terms. However, most of these methods approximate the $\ell_0$ norm with convex functions. To estimate more accurate solutions, it is desirable to treat the $\ell_0$ norm directly as a regularization term. In this study, we formulate the sparse estimation problem for Gaussian graphical models using the $\ell_0$ norm and propose a method to solve this problem using the Difference of Convex functions Algorithm (DCA). Specifically, we convert the $\ell_0$ norm constraint into an equivalent largest-$K$ norm constraint, reformulate the constrained problem into a penalized form, and solve it using the DC algorithm (DCA). Furthermore, we designed an algorithm that efficiently computes using graphical lasso. Experimental results with synthetic data show that our method yields results that are equivalent to or better than existing methods. Comparisons of model learning through cross-validation confirm that our method is particularly advantageous in selecting true edges.

\section*{Introduction}
\subsection*{Background}
Quantifying the relationships between variables as a structure from observed data is fundamental to data mining. One commonly used measure is Pearson's product-moment correlation coefficient, defined as the covariance of standardized variables. However, it has significant limitations, such as its inability to address spurious correlations. In contrast, the Gaussian Graphical Model (GGM) provides a method for learning partial correlations based on the precision matrix, which is the inverse of the covariance matrix. This approach graphically represents the relationships between variables while accounting for the influence of other variables. Such structural estimation is utilized across various fields, including estimating brain activity patterns \cite{ortiz2015exploratory}, anomaly detection \cite{ide2009proximity}, and sentiment analysis on social networks \cite{tan2011user}.

Since there are slight relationships between most variables, directly applying GGM often results in a dense graph with edges between all variables. Therefore, sparse estimation methods for GGM are actively researched to estimate more interpretable structures \cite{fan2016overview, drton2017structure, chen2023estimation}. Sparse estimation for GGM aims to represent the graph of relationships between variables in a simpler form by reducing the number of non-zero elements in the precision matrix. This approach allows for the estimation of interpretable graphs even when the number of variables is large or when the sample size is smaller than the number of variables. However, sparse estimation for GGM faces several technical challenges, including reducing computational complexity and ensuring the positive definiteness of the precision matrix.

\subsection*{Previous Research}
Sparse estimation methods for the precision matrix have long existed, including testing-based methods \cite{dempster1972covariance} and methods that select non-zero elements using thresholds \cite{bickel2008covariance}. A method that uses lasso, which adds the $\ell_1$ norm of the parameters to the loss function, proposed by Hyndman et al. \cite{tibshirani1996regression}, has also been suggested for sparse estimation \cite{meinshausen2006high, peng2009partial}. These methods face issues such as requiring extensive time for high-dimensional data or failing to ensure the positive definiteness of the precision matrix.

In this study, we focus on the method that has become mainstream in recent years, which involves adding regularization terms to the log-likelihood function. Banerjee et al. formulated the sparse structure estimation of GGM as a convex optimization problem by adding the $\ell_1$ norm of the precision matrix components to the log-likelihood function \cite{banerjee2006convex}. The graphical lasso method \cite{friedman2008sparse}, which solves this problem, is widely used because it works quickly and stably while ensuring positive definiteness, even when the number of variables is larger than the sample size or when correlations between variables are high. Graphical lasso is an iterative algorithm that solves the problem with the $\ell_1$ norm added to the log-likelihood function of GGM, determining the strength of sparsity by adjusting the penalty parameter. There are various derivative methods of graphical lasso \cite{cai2011constrained, rolfs2012iterative, mazumder2012graphical}.

Methods for determining the penalty parameter include information criteria, cross-validation, and methods focusing on the stability of the estimation results. Using information criteria such as AIC and BIC, the penalty parameter works well for a small number of variables but tends to estimate graphs with high false positive rates for high-dimensional data \cite{liu2010stability}. For BIC, using extended BIC makes it easier to reproduce the true graph compared to regular BIC when the number of true edges is small \cite{foygel2010extended, meinshausen2010stability}. Cross-validation estimates more true edges than using information criteria but has the problem of high model variability \cite{foygel2010extended}. Methods focusing on the stability of estimation results, such as subsampling, have shown improved accuracy in reproducing the true graph for high-dimensional data compared to methods using information criteria or cross-validation \cite{meinshausen2010stability, liu2010stability}. Recently proposed methods include minimizing a function of the network's characteristic index and penalty parameter with respect to the penalty parameter \cite{mestres2018selection}, and methods that extend to distributions other than the normal distribution to robustly estimate the precision matrix and covariance matrix \cite{avella2018robust, chun2018robust}.

While there are many successful cases of methods based on lasso, it is well known that estimators with the $\ell_1$ norm have biases. A desirable property for estimators, known as the oracle property \cite{fan2001variable}, has led to methods supplementing the shortcomings of lasso. Representative methods include SCAD \cite{fan2009network} and MCP \cite{zhang2010nearly}, which use continuous non-convex functions as regularization terms, and adaptive lasso \cite{zou2006adaptive}, which gives individual weights to the parameters. SELO \cite{dicker2013variable} has been designed with a regularization term that closely approximates the $\ell_0$ norm, which represents the number of non-zero elements, and a non-convex regularization term based on inverse trigonometric functions that converges to the $\ell_0$ norm has also been proposed \cite{wang2016variable}. However, both $\ell_1$ regularization and non-convex regularization approximate the $\ell_0$ norm with a more manageable function. Ideally, the $\ell_0$ norm should be introduced as a regularization term for sparse estimation. However, the $\ell_0$ norm is a non-convex and discontinuous function, making optimization difficult \cite{natarajan1995sparse}, and to our knowledge, there are no sparse estimation methods using $\ell_0$ regularization terms for GGM.

DC (Difference of Convex functions) optimization is one of a methods for solving sparse optimization problems with the $\ell_0$ norm \cite{neumann2005combined, le2015dc, gotoh2018dc}. This method expresses the non-convex objective function as the difference between two convex functions and repeatedly solves the convex optimization problem, which linearly approximates the concave function, to find a high-quality solution to the original non-convex optimization problem \cite{tao1986algorithms, tao1997convex}. Among them, the DC formulation proposed by Goto et al. \cite{gotoh2018dc} is unique in that it estimates a penalty parameter value that guarantees optimality for specific problems and does not use overly large constants, reporting favorable experimental results compared to lasso. DC optimization is applied to various classes, including quadratic programming and bilevel programming \cite{le2018dc}, and is actively researched for sparse estimation problems such as support vector machines \cite{neumann2005combined}.

\subsection*{Contributions}
In this study, we propose a sparse estimation method for GGM using the $\ell_0$ norm. As previously mentioned, sparse estimation with the $\ell_0$ norm is highly challenging, and to our knowledge, there are no existing studies that solve the problem of adding the $\ell_0$ norm to the log-likelihood function of GGM. We address this problem by applying the DC optimization method proposed by Goto et al. \cite{gotoh2018dc} to solve the problem constrained by the $\ell_0$ norm. Specifically, we first equivalently rewrite the constraint of the $\ell_0$ norm using the largest-$K$ norm defined by Goto et al. We then reformulate this constrained problem into a penalized form and introduce the DC formulation, which represents the unconstrained optimization problem as the difference of two convex functions. We design an algorithm to solve this problem and further utilize graphical lasso to achieve efficient solutions. 

The effectiveness of our method is validated using synthetic data. The data are generated using two methods, and for each, we conduct two experiments: one where a fixed number of edges are estimated, and the other where edges are estimated using cross-validation. The results show that our proposed method can generate the true graph with equal or better accuracy compared to existing methods in many settings. Particularly, our method delivers superior accuracy when estimating edges using cross-validation. Moreover, although our method repeatedly uses graphical lasso within the algorithm, the execution time is only a few times that of graphical lasso, confirming that the algorithm converges within a sufficiently practical timeframe.

\section*{Methods}
In this section, we first provide a detailed explanation of GGM and its common sparse estimation methods, as well as graphical lasso, which is closely related to our method. We then describe the sparse estimation of GGM using the DC algorithm, which is the core of our proposed method.

In this paper, we denote a consecutive set of positive integers as follows:
\[
[n] \coloneqq
\begin{cases}\notag
\{1,2,\ldots,n\} & \mbox{if $n \geq 1$}, \\
\emptyset & \mbox{otherwise}. \\
\end{cases}
\]

\subsection*{Sparse Estimation of Gaussian Graphical Models}
\subsubsection*{Gaussian Graphical Model}
A Gaussian graphical model (GGM) is a method for estimating and graphing the relationships between variables in a $p$-dimensional random vector $\bm{x} \coloneqq (x_1, x_2, \ldots, x_p)^{\top}$ that follows a multivariate normal distribution. Let $\mathcal{N}(\mu, \sigma^2)$ denote a normal distribution with mean $\mu$ and variance $\sigma^2$, and let $\omega_{jk}$ denote the $(j,k)$ element of the precision matrix $\bm{\Omega}$ (the inverse of the covariance matrix $\bm{\Sigma}$ of $\bm{x}$). The conditional distribution of $x_j$ given the other variables $\bm{x}_{-j} \coloneqq (x_k)_{k \neq j}$ can be written as follows:
\begin{align}\label{eq:cind_ggm}
\text{Pr}(x_j \mid \bm{x}_{-j}) \coloneqq \mathcal{N}\Bigl(-\frac{1}{\omega_{jj}^2}\sum_{k \neq j}{\omega_{jk}x_k},~\frac{1}{\omega_{jj}^2} \Bigr)
\end{align}
In this model, the relationship between $x_j$ and $x_k$ is determined by $\omega_{jk}$. Typically, maximum likelihood estimation is used to estimate the elements of the precision matrix. Given $n$ observed data points $\bm{x}_1, \bm{x}_2, \ldots, \bm{x}_n$, the sample mean vector $\hat{\bm{\mu}} \coloneqq \frac{1}{n}\sum_{i=1}^n{\bm{x}_i}$, and the sample covariance matrix $\bm{S} \coloneqq \frac{1}{n}\sum_{i=1}^n{(\bm{x}_i - \hat{\bm{\mu}})(\bm{x}_i - \hat{\bm{\mu}})^\top}$, the log-likelihood function with respect to $\bm{\Omega}$ can be written using the trace operator for square matrices $\text{tr}(\cdot)$ as follows:
\begin{align}
    \ell(\bm{\Omega}) &= \log \left( \Pi_{i=1}^{n}(2\pi)^{-\frac{p}{2}}|\bm{\Omega}|^{\frac{1}{2}}\exp \left[ -\frac{1}{2}(\bm{x}_i - \hat{\bm{\mu}})^\top\bm{\Omega}(\bm{x}_i - \hat{\bm{\mu}})\right] \right)\notag\\
    &= -\frac{pn}{2}\log(2\pi) + \frac{n}{2}\log |\bm{\Omega}| - \frac{1}{2}\sum_{i=1}^n{(\bm{x}_i - \hat{\bm{\mu}})^\top\bm{\Omega}(\bm{x}_i - \hat{\bm{\mu}})} \notag\\
    &= -\frac{pn}{2}\log(2\pi) + \frac{n}{2}\log |\bm{\Omega}| - \frac{n}{2}\text{tr}(\bm{\Omega}\bm{S}).\notag
\end{align}
Here, we used $\bm{x}^\top\bm{\Omega}\bm{x} = \text{tr}(\bm{\Omega}\bm{x}\bm{x}^\top)$. Furthermore, by removing the constant terms and coefficients that are irrelevant to the optimization and multiplying by negative to convert it to a minimization problem, we obtain:
\begin{align}\label{eq:loglike_ggm}
- \text{log}|\bm{\Omega}| + \text{tr}(\bm{\Omega} \bm{S})
\end{align}
The maximum likelihood estimator of the precision matrix is given by $\hat{\bm{\Omega}} = \bm{S}^{-1}$.

If $\omega_{jk}=0~(j \neq k)$ in Eq~\eqref{eq:cind_ggm}, $x_k$ does not influence $x_j$ given $\bm{x}_{-j}$, which is called conditional independence. By estimating $\omega_{jk}$ as exactly zero, we can estimate a sparse graph by connecting only the variables that are not conditionally independent. To sparsely estimate the precision matrix, we add a regularization term $p_{\lambda}(\bm{\Omega})$ to Eq~\eqref{eq:loglike_ggm} to penalize the values of the elements of the precision matrix:
\begin{align}\label{eq:loglike_penalty}
- \text{log}|\bm{\Omega}| + \text{tr}(\bm{\Omega} \bm{S}) + p_{\lambda}(\bm \Omega).
\end{align}
Here, $\lambda>0$ is the penalty parameter that determines the strength of the penalty imposed by $p_\lambda(\cdot)$. A larger $\lambda$ imposes a stronger penalty, resulting in more parameters being estimated as zero. Various types of sparse estimation can be represented by the choice of the regularization term $p_{\lambda}(\bm{\Omega})$. For example, setting $p_{\lambda}(\bm \Omega) \coloneqq \lambda\|\text{vec}(\bm{\Omega})\|_1$ corresponds to graphical lasso\cite{friedman2008sparse}. For a given input $x \in \mathbb{R}$, the SCAD \cite{fan2009network} regularization term can be expressed using a penalty parameter $\lambda$ and another parameter $a>0$ as follows:
\begin{align}
\text{SCAD}_{\lambda, a}(x) &= \begin{cases} 
\lambda |x| & \text{if } |x| \leq \lambda, \\
\frac{a\lambda |x| - \frac{x^2 + \lambda^2}{2}}{a - 1} & \text{if } \lambda < |x| \leq a\lambda, \\
\frac{(a + 1)\lambda^2}{2} & \text{if } |x| > a\lambda.
\end{cases} \label{eq:scad}
\end{align}
Using this function, the objective function with the SCAD regularization term can be formulated as:
\begin{align*}
    - \text{log}|\bm{\Omega}| + \text{tr}(\bm{\Omega} \bm{S}) + \sum_{j=1}^p\sum_{k=1}^p{\text{SCAD}_{\lambda, a}(\omega_{jk})}
\end{align*}
Additionally, by using adaptive lasso\cite{zou2006adaptive}, a weighted version of lasso, the objective function can be written as:
\begin{align}\label{eq:adapt}
    - \text{log}|\bm{\Omega}| + \text{tr}(\bm{\Omega} \bm{S}) + \lambda\sum_{j=1}^p\sum_{k=1}^p{\frac{1}{|\Tilde{\omega}_{jk}|^\gamma}|\omega_{jk}|}, 
\end{align}
where $\bm{\tilde{\Omega}}:=(\tilde{\omega}_{jk})_{(j,k)\in[p]\times[p]}$ is a consistent estimator of $\bm{\Omega}$ and $\gamma>0$ is a parameter. Fig~\ref{Fig:graph_ex} illustrates the characteristics of the regularization terms of graphical lasso (glasso), SCAD, and adaptive lasso with parameters $\lambda=0.5$, $a=3.7$, and $\gamma=0.5$.
\begin{figure}[ht]
\center
\includegraphics[width=\linewidth]{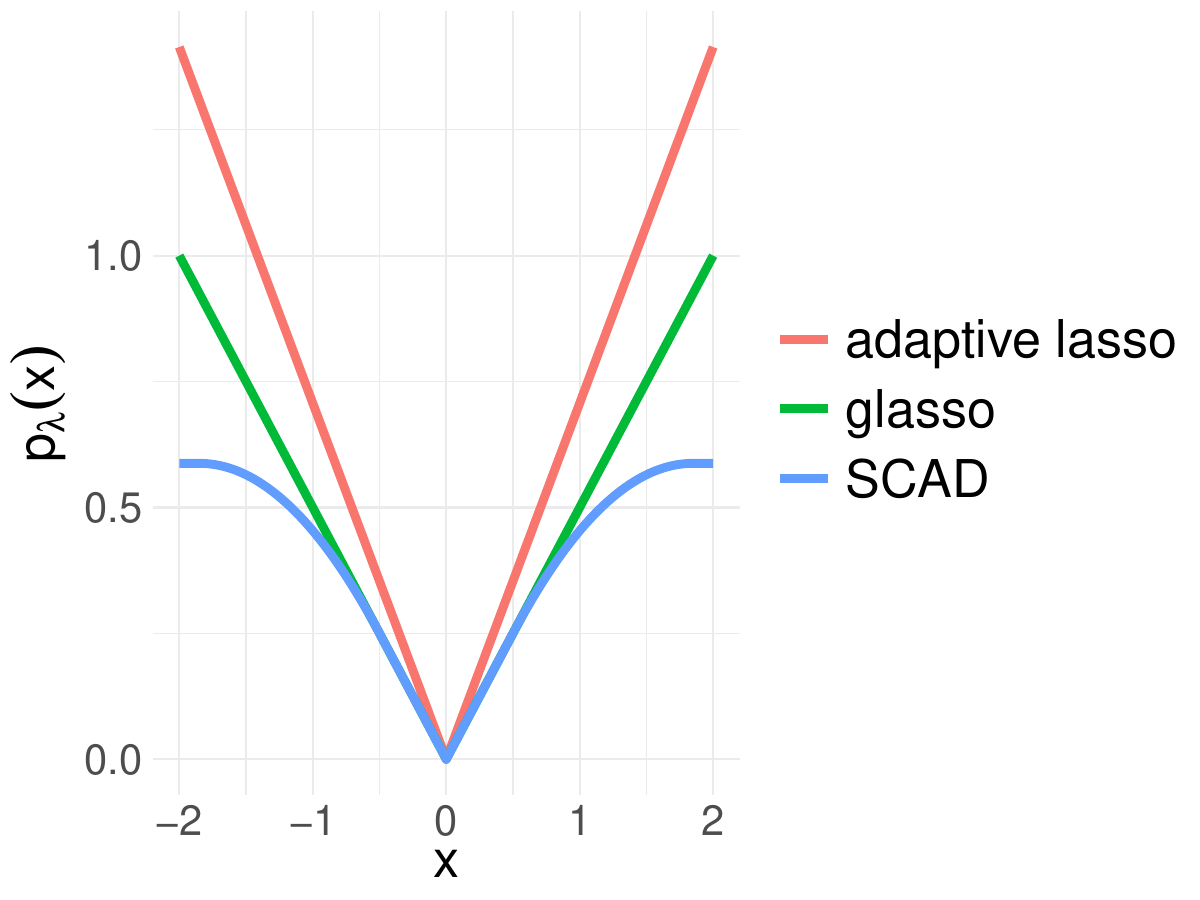}
\caption{Penalty values by each method}
\label{Fig
}
\end{figure}

\subsubsection*{Graphical Lasso}
Graphical lasso\cite{friedman2008sparse}, which is deeply involved in the algorithm of our proposed method, uses the regularization term $p_{\lambda}(\bm \Omega) \coloneqq \lambda\|\text{vec}(\bm \Omega)\|_1$ and considers the following nonlinear equation obtained by differentiating Eq~\eqref{eq:loglike_penalty} with respect to $\bm{\Omega}$ and setting it to zero:
\begin{equation}\label{eq:eq_loglike_glasso}
-\bm \Omega^{-1} + \bm S + \lambda \bm \Gamma(\bm{\Omega}) = \bm O
\end{equation}
where
\begin{gather}
\bm \Gamma(\bm{\Omega}) \coloneqq (\gamma_{jk}(\omega_{jk}))_{(j,k) \in [p] \times [p]} \in \mathbb{R}^{p \times p}, \label{eq:Gamma}\\
\gamma_{jk}(\omega_{jk}) \in \text{sign}^*(\omega_{jk}) \coloneqq 
\begin{cases}\notag
\{\text{sign}(\omega_{jk})\} & \mbox{if $\omega_{jk} \neq 0$}, \\
[-1,1] & \mbox{if $\omega_{jk} = 0$}. \\
\end{cases}
\end{gather}
The function $\text{sign}(x)$, called the sign function, is defined as follows:
\[
\text{sign}(x) = 
\begin{cases}\notag
1 & \mbox{if $x > 0$}, \\
0 & \mbox{if $x=0$}, \\
-1 & \mbox{if $x<0$}
\end{cases}
\]
Eq~\eqref{eq:eq_loglike_glasso} has $p \times p$ components, but this problem can be effectively solved by introducing the following block matrices and focusing on the $i$-th row and $i$-th column:
\begin{align}\label{eq:glasso_matrix_separated}
\bm{\Omega} = 
\begin{bmatrix}
\bm{\Omega}_{-i} & \bm{\omega}_i \\
\bm{\omega}_i^\top & \omega_{ii}
\end{bmatrix}, \quad 
\bm{\Sigma} = 
\begin{bmatrix}
\bm{\Sigma}_{-i} & \bm{\sigma}_i \\
\bm{\sigma}_i^\top & \sigma_{ii} \\
\end{bmatrix}, \quad
\bm{S} = 
\begin{bmatrix}
\bm{S}_{-i} & \bm{s}_i \\
\bm{s}_i^\top & s_{ii}
\end{bmatrix}.
\end{align}
Here, $\bm{\Omega}_{-i}, ~\bm{\Sigma}_{-i}, ~\bm{S}_{-i}$ represent the $(p-1) \times (p-1)$ matrices excluding the $i$-th row and column, and $\bm{\omega}_i, ~\bm{\sigma}_i, ~\bm{s}_i$ are the vectors excluding the diagonal elements $\omega_{ii}, ~\sigma_{ii}, ~s_{ii}$ from the $i$-th column vectors. This problem reduces to lasso regression\cite{tibshirani1996regression} for each row and can be solved quickly using coordinate descent\cite{friedman2008sparse}.

The estimation procedure of graphical lasso is summarized in Algorithm~\ref{alg:glasso_alg}. In the initialization, the covariance matrix $\bm{\Sigma}_0 = \bm{S} + \lambda \bm{I}_p$ is obtained using the diagonal elements determined from Eq~\eqref{eq:eq_loglike_glasso} and the off-diagonal elements obtained by maximum likelihood estimation. Note that this algorithm has been criticized for the objective function not decreasing monotonically, and methods like P-GLASSO and DP-GLASSO have been proposed to accelerate convergence\cite{mazumder2012graphical}.

\begin{algorithm}
\caption{Graphical Lasso Algorithm}
\label{alg:glasso_alg}
\textbf{Input:} Sample covariance matrix $\bm{S}$, penalty parameter $\lambda$ \\
\textbf{Output:} Precision matrix $\bm{\Omega}$ \\
\textbf{Initialize:} Covariance matrix $\bm{\Sigma}_0 = \bm{S} + \lambda \bm{I}_p$, precision matrix $\bm{\Omega}_0 = \bm{\Sigma}_0^{-1}$, threshold $\varepsilon>0$ for convergence.
\begin{algorithmic}[1]
\WHILE{$-\log |\bm{\Omega}| + \text{tr}(\bm{\Omega}\bm{S}) + \lambda\|\text{vec}(\bm{\Omega})\|_1 \geq \varepsilon$}
\FOR{$i \in [p]$}
\STATE Rearrange into block matrices by moving the $i$-th row and column as in Eq~\eqref{eq:glasso_matrix_separated}
\STATE Update $\omega_{ii}, \bm{\omega}_i, \sigma_{ii}, \bm{\sigma}_i$ by lasso regression\cite{friedman2008sparse}
\STATE Rearrange the elements back into the original matrix
\ENDFOR
\ENDWHILE
\RETURN $\bm{\Omega}$
\end{algorithmic}
\end{algorithm}

\subsection*{Proposed Method: Sparse Estimation of GGM via DC Algorithm}
\subsubsection*{Formulation}
Sparse estimation for GGM can be described as the following cardinality-constrained optimization problem using the cardinality parameter $K \in [p^2]$.
\begin{align}\label{eq:original_l0}
\text{minimize} \quad& -\text{log}|\bm{\Omega}| + \text{tr}(\bm{\Omega} \bm{S}) \\ \notag
\text{subject to} \quad& \|\text{vec}(\bm{\Omega})\|_0 \leq K.
\end{align}
Here, $\|\text{vec}(\bm{\Omega})\|_0 \coloneqq | \{ (j,k) \in [p] \times [p] \mid \omega_{jk} \neq 0 \} |$ is called the $\ell_0$ norm. Following Goto et al.\cite{gotoh2018dc}, we define the largest-$K$ norm as follows.
\begin{dfn}
For $\bm{w} \coloneqq (w_i)_{i \in [n]} \in \mathbb{R}^n$ with $|w_{(1)}| \ge |w_{(2)}| \ge \cdots \ge |w_{(n)}|$, the largest-$K$ norm $\opnorm{\bm{w}}_K \in \mathbb{R}$ represents the sum of the absolute values of the largest $K$ components of $\bm{w}$. That is,
\begin{align}
\opnorm{\bm{w}}_K := \sum^K_{i=1}|w_{(i)}|.
\end{align}
\end{dfn}

The constraint $\|\bm{w}\|_0$ in Eq~\eqref{eq:original_l0} represents that the number of non-zero components of $\bm{w}$ is at most $K$. This is equivalent to $\sum_{i=K+1}^n{w_{(i)}}=0$, which further translates to $\sum_{i=K+1}^n{w_{(i)}} = \|\bm{w}\|_1 - \opnorm{\bm{w}}_K=0$. Thus, using the largest-$K$ norm, Eq~\eqref{eq:original_l0} can be equivalently rewritten as:
\begin{align}\label{eq:dc_constrained}
\text{minimize} \quad& - \text{log}|\bm{\Omega}| + \text{tr}(\bm{\Omega} \bm{S}) \\ \notag
\text{subject to} \quad& \|\text{vec}(\bm{\Omega})\|_1 - \opnorm{\text{vec}(\bm{\Omega})}_K=0.
\end{align}

This constraint is called a DC constraint since it is represented as the difference of two convex functions. The $\ell_0$ norm constraint $\|\text{vec}(\bm{\Omega})\|_0$ in Eq~\eqref{eq:original_l0} is a discontinuous function, while the continuous function $\|\text{vec}(\bm{\Omega})\|_1 - \opnorm{\text{vec}(\bm{\Omega})}_K$ in Eq~\eqref{eq:dc_constrained} represents the same feasible region as the original problem.

To solve Eq~\eqref{eq:dc_constrained}, we reformulate it into a penalized form using a penalty parameter $\eta>0$ as follows:
\begin{align}
\text{minimize} \quad &-\text{log}|\bm{\Omega}| + \text{tr}(\bm{\Omega} \bm{S}) + \eta(\|\text{vec}(\bm{\Omega})\|_1 - \opnorm{\text{vec}(\bm{\Omega})}_K)\notag \\
&= (-\text{log}|\bm{\Omega}| + \text{tr}(\bm{\Omega} \bm{S}) + \eta \|\text{vec}(\bm{\Omega})\|_1) - \eta \opnorm{\text{vec}(\bm{\Omega})}_K.\label{eq:dc_penalty}
\end{align}

Here, $-\text{log}|\bm{\Omega}| + \text{tr}(\bm{\Omega}\bm{S}) + \eta \|\text{vec}(\bm{\Omega})\|_1$ and $\eta \opnorm{\text{vec}(\bm{\Omega})}_K$ are convex functions. Similar to the previous DC constraint, optimization for such a problem where the objective function is represented as the difference of two convex functions is called DC optimization\cite{tao1986algorithms}.

\subsubsection*{Algorithm}\label{sec:proposition_alg}
In DC optimization, the subdifferential $\partial \opnorm{\text{vec}(\bm{\Omega}_t)}_K$ at the provisional solution $\bm{\Omega}_t$ at time $t$ is used to linearly approximate the largest-$K$ norm $\opnorm{\text{vec}(\bm{\Omega})}_K$ in Eq~\eqref{eq:dc_penalty}. The subgradient $\bm{s}(\text{vec}(\bm{\Omega}_t))$ at time $t$ is obtained as follows:
\begin{align}\label{eq:dc_step1}
\bm{s}(\text{vec}(\bm{\Omega}_t)) &\in  \partial\opnorm{\text{vec}(\bm{\Omega}_t)}_K \\
&= \underset{\bm{s}}{\text{argmax}}\{\text{vec}(\bm{\Omega}_t)^\top \bm{s} \mid \|\bm{s}\|_1=K, ~|s_i| \leq 1, ~i \in [p^2]\}
\end{align}
This problem can be solved by applying the sign function to $\text{vec}(\bm{\Omega}_t)_{(1)}, \text{vec}(\bm{\Omega}_t)_{(2)}, \ldots, \text{vec}(\bm{\Omega}_t)_{(K)}$, and setting the other components to 0.

Using $\bm{s}(\text{vec}(\bm{\Omega}_t))$, the linear approximation of the objective function in Eq~\eqref{eq:dc_penalty} is obtained as follows:
\begin{align}\label{eq:dc_step2}
g_t(\bm{\Omega}) :=  -\text{log}|\bm{\Omega}| + \text{tr}(\bm{\Omega}\bm{S}) + \eta \|\text{vec}(\bm{\Omega})\|_1 - \eta\text{vec}(\bm{\Omega})^\top \bm{s}(\text{vec}(\bm{\Omega}_t))
\end{align}
To minimize this, $g_t$ is differentiated with respect to $\bm{\Omega}$, and the components arranged as a vector are rearranged into the original matrix, giving the following equation:
\begin{align}\label{eq:dc_glasso}
\frac{\partial g_t}{\partial \bm{\Omega}} = -\bm{\Omega}^{-1} + (\bm{S} - \eta \bm{V}(\bm{\Omega}_t)) + \eta \bm{\Gamma}(\bm{\Omega}) = 0
\end{align}
Here, $\bm{\Gamma}(\bm{\Omega})$ is as in Eq~\eqref{eq:Gamma}, and $\bm{V}(\bm{\Omega}_t) \in \mathbb{R}^{p\times p}$ is a matrix obtained by rearranging the components of the subgradient $\bm{s}(\text{vec}(\bm{\Omega}_t))$ as $\bm{V}(\bm{\Omega}_t) \coloneqq \text{vec}^{-1}(\bm{s}(\bm \Omega_t))$. This is a nonlinear equation where $\bm{S}$ in Eq~\eqref{eq:eq_loglike_glasso} is replaced by $\bm{S}-\eta \bm{V}(\bm{\Omega}_t)$, and the graphical lasso algorithm can be directly applied.

The graphical lasso estimation algorithm assumes that the sample covariance matrix $\bm{S}$ is positive definite. However, in Eq~\eqref{eq:dc_glasso}, the corresponding sample covariance matrix is $\bm{S} - \eta\bm{V}(\bm{\Omega}_t)$, which may not be positive definite depending on the value of $\eta$. Therefore, $\eta$ must be chosen from a set of values that make $\bm{S} - \eta\bm{V}(\bm{\Omega}_t)$ positive definite. Since the diagonal elements of $\bm{V}(\bm{\Omega}_t)$ are always 1 (due to the positive definiteness of the precision matrix), setting $\eta$ to the minimum value of the diagonal elements of $\bm{S}$ ensures that the minimum eigenvalue of $\bm{S} - \eta\bm{V}(\bm{\Omega}_t)$ is 0, making it not positive definite. Based on this fact, $\eta$ is initially set to the minimum diagonal element of $\bm{S}$ and then decreased by a constant rate $\alpha$ until $\bm{S}-\eta\bm{V}(\bm{\Omega}_t)$ becomes positive definite (i.e., the minimum eigenvalue is positive). In this study, we set $\alpha = 0.5$.

The proposed sparse estimation method for the precision matrix is summarized in Algorithm~\ref{alg:dcglasso_alg}. In this study, $\bm{\Omega}_0 = (\bm{S} + \bm{I}_p)^{-1}$ was used.

\begin{algorithm}
\caption{DC Algorithm for GGM}
\label{alg:dcglasso_alg}
\textbf{Input:} Sample covariance matrix $\bm{S}$, cardinality parameter $K\in[p^2]$\\
\textbf{Output:} Precision matrix $\bm{\Omega}$\\
\textbf{Initialize:} Iteration step $t \leftarrow 0$, convergence condition $\varepsilon>0$, positive definite matrix $\bm{\Omega}_0$, step width $\alpha$ for $\eta_t$
\begin{algorithmic}[1]
\REPEAT
    \STATE Solve for the subgradient $\bm{s}(\text{vec}(\bm{\Omega}_t)) \in \partial\opnorm{\text{vec}(\bm{\Omega}_t)}_K$ and compute $\bm{s}(\text{vec}(\bm{\Omega}_t))$.
    \STATE Set $\eta_t$ such that $\bm{S} - \eta_t\bm{V}(\bm{\Omega}_t)$ is positive definite using the step width $\alpha$.
    \STATE Solve Eq~\eqref{eq:dc_glasso} using Algorithm~\ref{alg:glasso_alg} and compute $\bm{\Omega}_{t+1}$.
    \STATE $t \leftarrow t + 1$
\UNTIL{$||\bm{\Omega}_{t} - \bm{\Omega}_{t-1}||_F^2 < \varepsilon$}
\RETURN $\bm{\Omega}_t$
\end{algorithmic}
\end{algorithm}

\section*{Experimental Results}
In this section, we conduct two experiments on synthetic data generated by two different methods to validate the effectiveness of our method: one method involves cross-validation for edge estimation, and the other estimates a fixed number of edges.

\subsection*{Experiments Using Synthetic Data}
We prepared synthetic data by two methods: random and chain. In each case, we design the true precision matrix $\bm{\Omega}$ and generate independent samples from a multivariate normal distribution using the true covariance matrix $\bm{\Sigma}=\bm{\Omega}^{-1}$. To ensure the definiteness of the matrices, the sample covariance matrix is shrunk as described in \cite{touloumis2015nonparametric}. The details of the precision matrix generation methods are as follows.

\begin{description}
\item[Random]
First, create a symmetric matrix $\bm{A}_2 := 0.5(\bm{A}_1 + \bm{A}_1^\top)$ using $\bm{A}_1 \in \mathbb{R}^{p\times p}$, whose components are independently generated as normal random numbers. To introduce sparsity, randomly set the off-diagonal components of $\bm{A}_2$ to zero while maintaining symmetry. Compute $\bm{\Omega}_\text{rnd}$ as $\bm{A}_2 + \eta \bm{I}_p$, where $\eta$ is chosen such that the minimum eigenvalue of $\bm{\Omega}_\text{rnd}$ is 1.

\item[Chain]
Set up a tridiagonal matrix as follows:
\[
\omega_{jk} =
\begin{cases}\notag
1 & \mbox{if $j = k$}, \quad (j,k) \in [p] \times [p], \\
0.5 & \mbox{if $|j - k| = 1$}, \quad (j,k) \in [p] \times [p], \\
0.25 & \mbox{if $|j - k| = 2$}, \quad (j,k) \in [p] \times [p], \\
0 & \mbox{otherwise}.
\end{cases}
\]
Randomly set some of the non-zero off-diagonal components to zero while maintaining symmetry to obtain the true precision matrix $\bm{\Omega}_\text{chn}$.
\end{description}

\begin{figure}[ht]
\centering
\includegraphics[width=\linewidth]{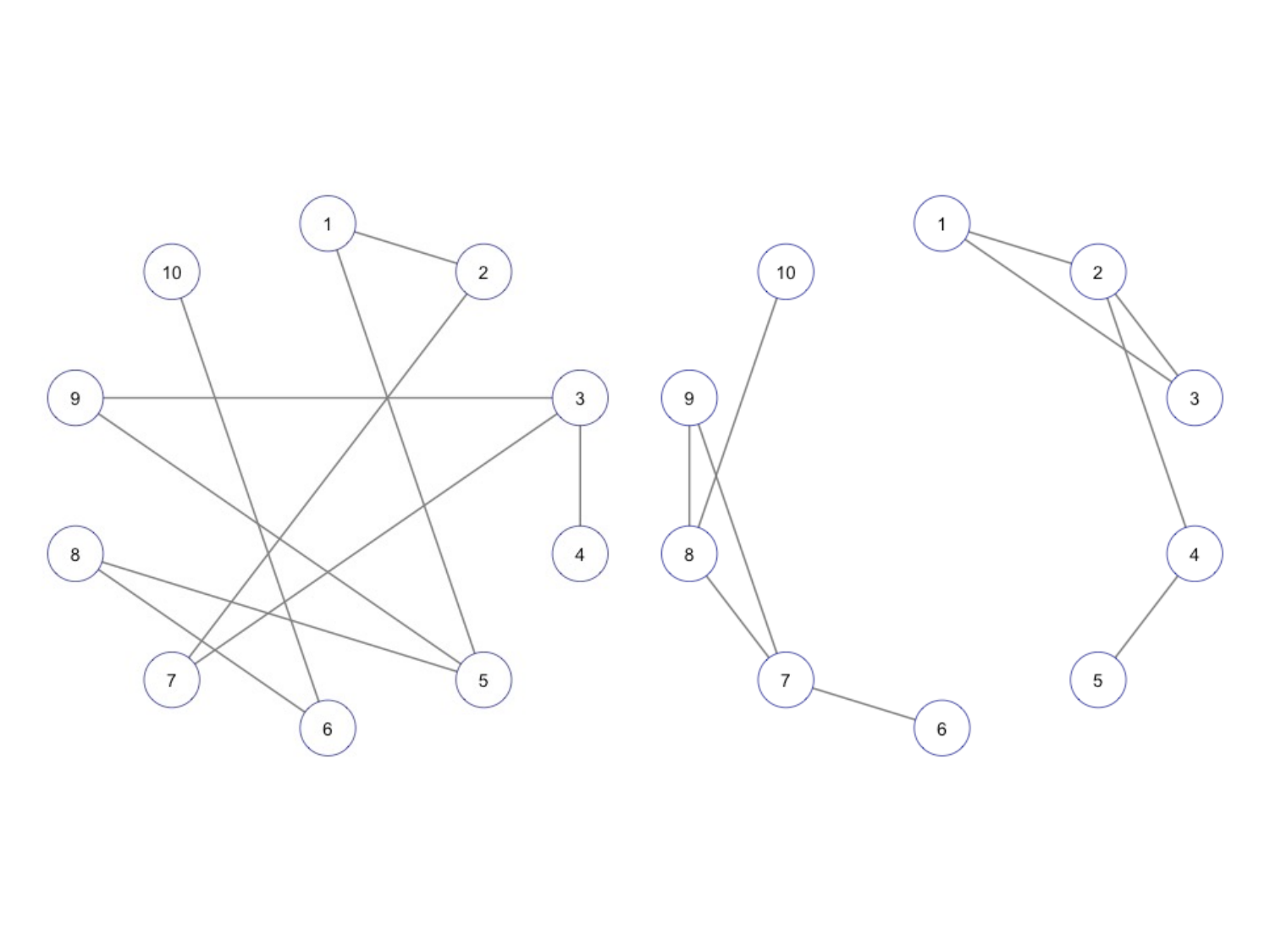}
\caption{\bf Examples of true graphs with $p=10$ and 10 non-zero components. Left: Random, Right: Chain}
\label{Fig:graph_ex}
\end{figure}

Fig~\ref{Fig:graph_ex} shows examples of graphs based on the precision matrices generated by each method. In this study, the number of non-zero off-diagonal components of the precision matrix $n_{\neq 0}$ was set to 30 for all settings. The procedure for creating synthetic data is summarized in Algorithm~\ref{alg: make_synthetic}.

\begin{algorithm}
\caption{Synthetic Data Generation Method}
\label{alg: make_synthetic}
\begin{algorithmic}[1]
\STATE \textbf{Input:} Sample size $n$, number of variables $p$, number of non-zero off-diagonal components $n_{\neq 0}$
\STATE \textbf{Output:} Sample covariance matrix $\bm{S}$
\STATE Generate a precision matrix $\bm{\Omega}$ with $n_{\neq 0}$ non-zero off-diagonal components, and compute the covariance matrix $\bm{\Sigma}=\bm{\Omega}^{-1}$.
\STATE Independently generate $n$ samples $\bm{x}_i \in \mathbb{R}^p~(i \in [n])$ from a multivariate normal distribution with mean vector $\bm{0}$ and covariance matrix $\bm{\Sigma}$, and compute the sample covariance matrix $\bm{S}$.
\STATE Compute the optimal shrinkage parameter $\zeta$ and update $\bm{S}$ as follows, where $\bm{D}_{\bm{S}}$ is a diagonal matrix with the diagonal elements of $\bm{S}$ and zero off-diagonal elements \cite{touloumis2015nonparametric}.\
$\bm{S} \leftarrow \zeta\bm{D}_{\bm{S}} + (1 - \zeta)\bm{S}$
\RETURN $\bm{S}$
\end{algorithmic}
\end{algorithm}

Two types of experiments are performed on each dataset to compare accuracy. Due to the randomness in the precision matrix generation method, 30 precision matrices are created for each setting. We compare the results using the average F1 scores and confidence intervals of the estimations.

\begin{description}
\item[Experiment 1: Estimating the Number of Edges Using Cross-Validation] \leavevmode \\
The final graph is estimated by models trained using 5-fold cross-validation. We use error function defined in Eq~\eqref{eq:loglike_ggm} during cross-validation. The search range for the parameter $K$ of our method, representing the number of non-zero off-diagonal edges, is set from $p+1$ (indicating one or fewer non-zero edges) to $\frac{p(p+1)}{2}$ (allowing all edges to be non-zero) with 100 evenly spaced points. For the penalty parameter $\lambda$ of existing methods, $\lambda_{\text{max}}$ is chosen such that the number of edges becomes zero for the sample covariance matrix $\bm{S}$, and 100 evenly spaced points are extracted from the range $[0,~\lambda_{\text{max}}]$.

\item[Experiment 2: Estimating a Fixed Number of Edges] \leavevmode \\
The accuracy is compared by adjusting the parameter $K$ of our method and the penalty parameter $\lambda$ of existing methods so that the number of non-zero edges estimated by each model is the same. The number of edges estimated is set to 20, 30, and 40, common to all settings.
\end{description}

Additionally, synthetic data is generated with the number of variables $p$ and sample size $n$ in the following 12 combinations, and similar experiments are conducted for each:
\[
p \in \{50, ~100, ~200, ~400\}, ~n \in \{p/2, ~p, ~2p\}.
\]

\subsection*{Methods for Comparison and Evaluation Metrics}
To validate the effectiveness of our method, we compare the accuracy and characteristics of the estimation results using the following methods:
\begin{description}
\item[DC:] Our DC algorithm (Algorithm~\ref{alg:dcglasso_alg}).
\item[glasso:] Graphical lasso (Algorithm~\ref{alg:glasso_alg}) \cite{friedman2008sparse}.
\item[SCAD:] SCAD regularization method \cite{fan2009network}. The parameter $a$ in Eq~\eqref{eq:scad} is set to $a=3.7$ as recommended in the original paper \cite{fan2001variable}.
\item[adapt:] Adaptive lasso \cite{zou2006adaptive}. The parameter $\gamma$ in Eq~\eqref{eq:adapt} is set to $\gamma = 0.5$ as specified in \cite{fan2009network}.
\end{description}

All experiments were conducted using the R programming language. For glasso, including those executed internally within DC, we used the \texttt{glasso} package \cite{friedman2008sparse}. For SCAD and adapt, we used the \texttt{GGMncv} package \cite{williams2020beyond}. The source code to reproduce these experimental results is available at "https://github.com/torikaze/DC-GGM".

The goodness of fit between the graph based on the estimated precision matrix $\hat{\bm{\Omega}} \coloneqq (\hat{\omega}_{jk})_{(j,k) \in [p] \times [p]} \in \mathbb{R}^{p \times p}$ and the true graph is evaluated using the F1 score. The F1 score based on the precision matrix is defined as follows:

\begin{gather} \notag
\text{F1 score} \coloneqq \frac{2 \times \text{precision} \times \text{recall}}{\text{precision} + \text{recall}}. \notag
\end{gather}
Here, $\text{precision} = \text{TP} / (\text{TP} + \text{FP})$ and $\text{recall} = \text{TP} / (\text{TP} + \text{FN})$, where TP (True Positive), FP (False Positive), and FN (False Negative) are defined as follows:
\begin{gather}
\text{TP}=\Sigma_{j<k}{\text{I}(\hat{\omega}_{jk} \neq 0 \quad \text{and} \quad \omega_{jk} \neq 0)}, \notag \\
\text{FP}=\Sigma_{j<k}{\text{I}(\hat{\omega}_{jk} \neq 0 \quad \text{and} \quad \omega_{jk} = 0)}, \notag \\
\text{FN}=\Sigma_{j<k}{\text{I}(\hat{\omega}_{jk} = 0  \quad \text{and} \quad \omega_{jk} \neq 0)} \notag.
\end{gather}
Here, $\text{I}(\text{cond})$ is an indicator function that returns $1$ if $\text{cond}$ is true and $0$ otherwise.

\subsection*{Experiment 1: Estimating the Number of Edges Using Cross-Validation}
\begin{figure}[ht]
\centering
\includegraphics[width=\linewidth]{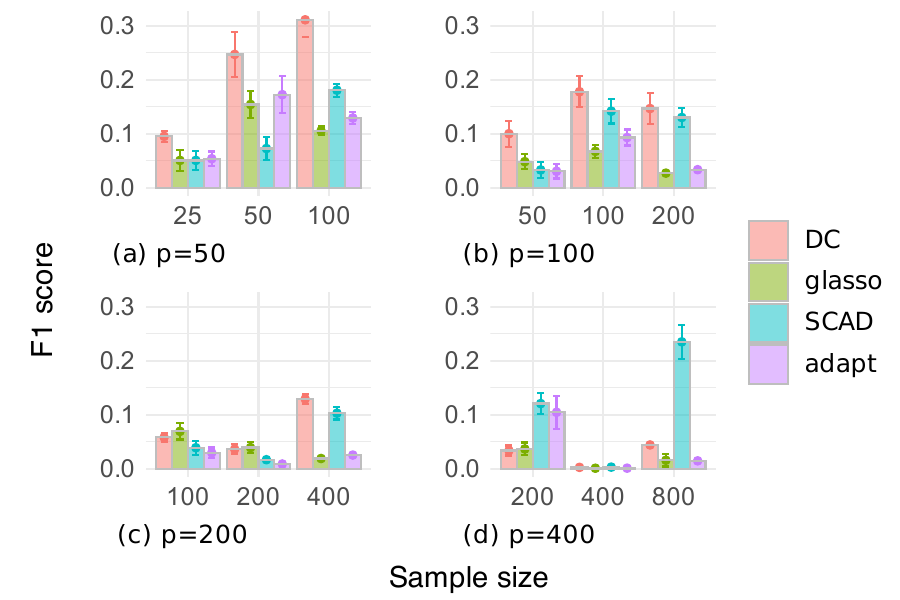}
\caption{\bf Results of edge estimation using cross-validation for random data}
\label{Fig:exp1_random}
\end{figure}

\begin{figure}[ht]
\centering
\includegraphics[width=\linewidth]{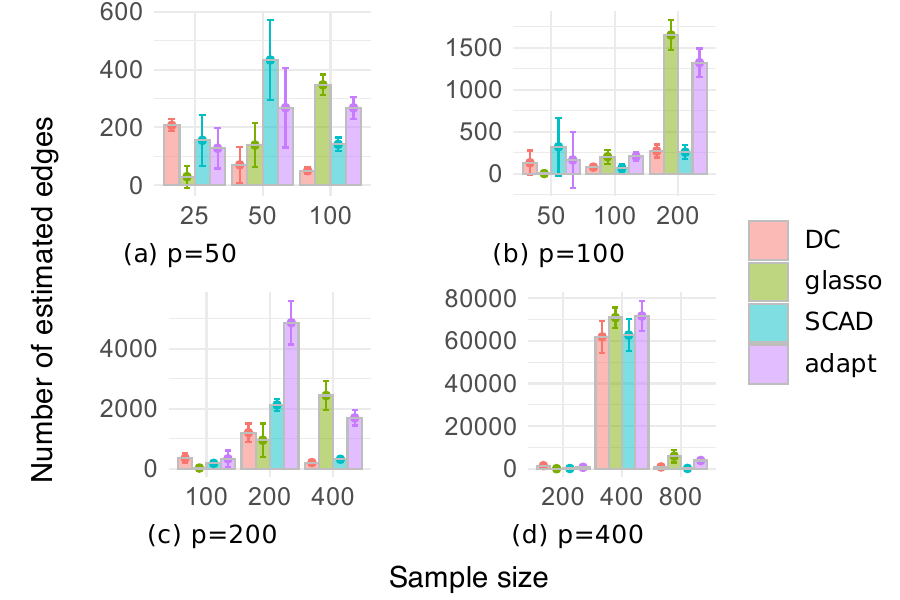}
\caption{\bf Number of edges selected through cross-validation for random data}
\label{Fig:exp1_random_numedge}
\end{figure}

Figures~\ref{Fig:exp1_random} and~\ref{Fig:exp1_random_numedge} show the results of Experiment 1 on random data, representing the F1 score and the number of estimated non-zero edges, respectively. Our method generally outperforms the others, except when $p=400$. The accuracy of our method (DC) improves as the sample size increases. Fig~\ref{Fig:exp1_random_numedge} indicates that glasso, SCAD and adapt estimate a significantly higher number of non-zero edges, leading to lower F1 scores. In contrast, glasso tends to estimate overly sparse graphs when $n < p$, resulting in lower F1 scores. DC delivered relatively small variance in the number of estimated edges, indicating robust estimation against data changes.

\begin{figure}[ht]
\centering
\includegraphics[width=\linewidth]{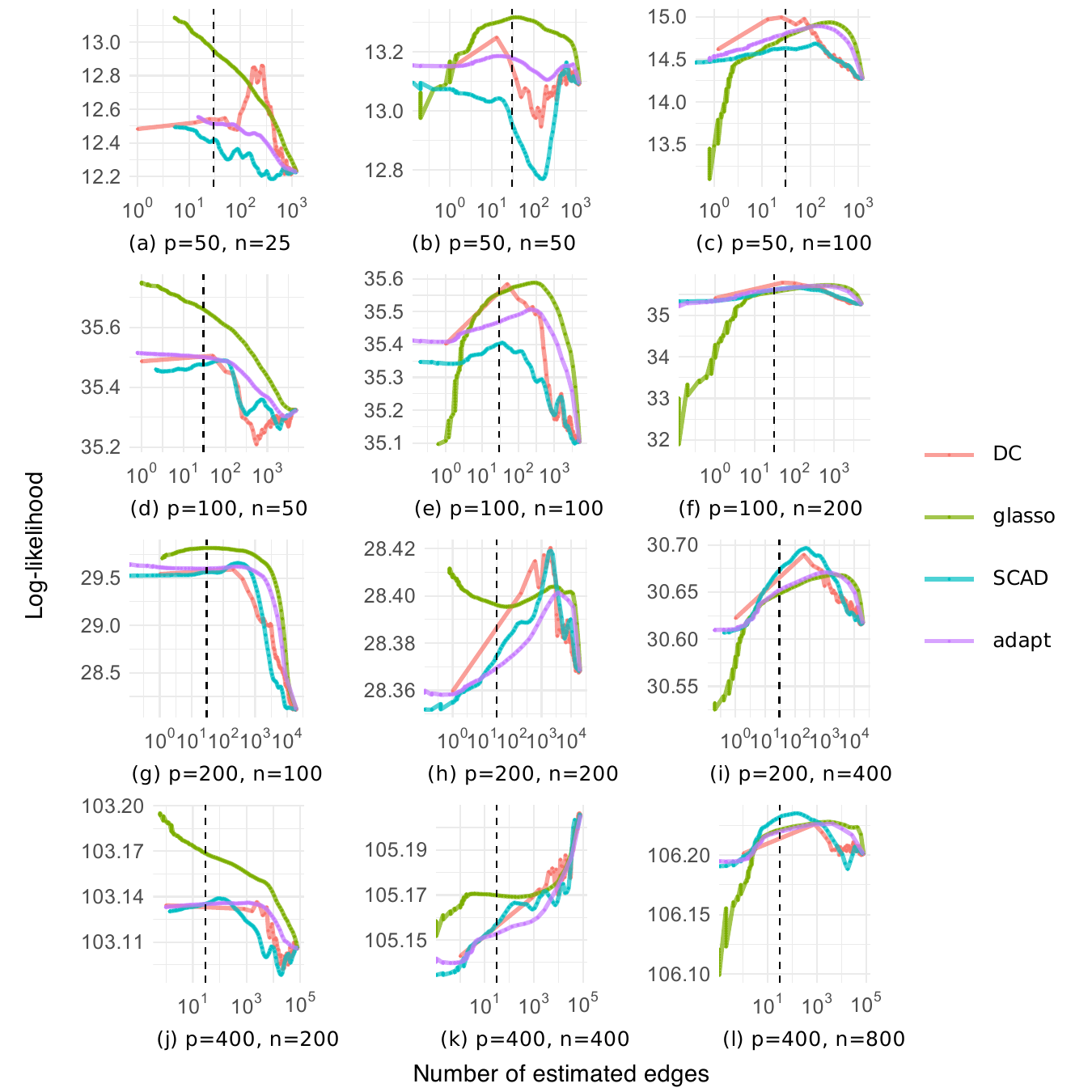}
\caption{\bf Number of selected edges and log-likelihood for random data (black dotted line: true number of edges)}
\label{Fig:exp6_random}
\end{figure}

To verify the number of edges selected through cross-validation, the relationship between the average number of selected edges and the average log-likelihood in cross-validation is shown in Fig~\ref{Fig:exp6_random}. Note that in Experiment 1, data was generated for 30 patterns by changing the initial random seed, but this figure shows the result for a single data set. As a general trend, stronger penalties (fewer selected edges) are chosen when $p > n$, while weaker penalties (more selected edges) are chosen when $p < n$. DC often maximizes the log-likelihood closer to the true number of edges compared to other methods. However, the relationship between the number of selected edges and the log-likelihood is not as smooth as it is with other methods.

\begin{figure}[ht]
\centering
\includegraphics[width=\linewidth]{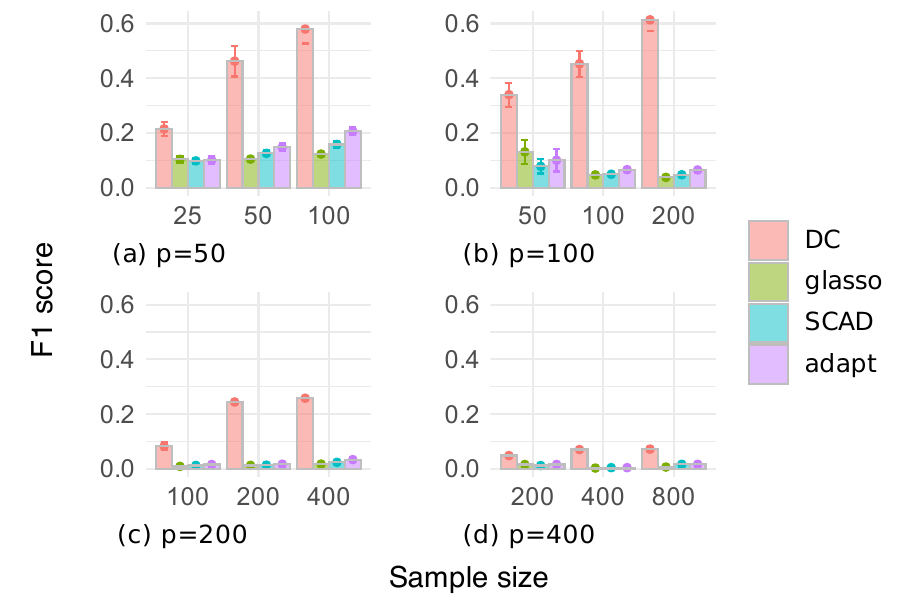}
\caption{\bf Results of edge estimation using cross-validation for chain data}
\label{Fig:exp1_chain}
\end{figure}

\begin{figure}[ht]
\centering
\includegraphics[width=\linewidth]{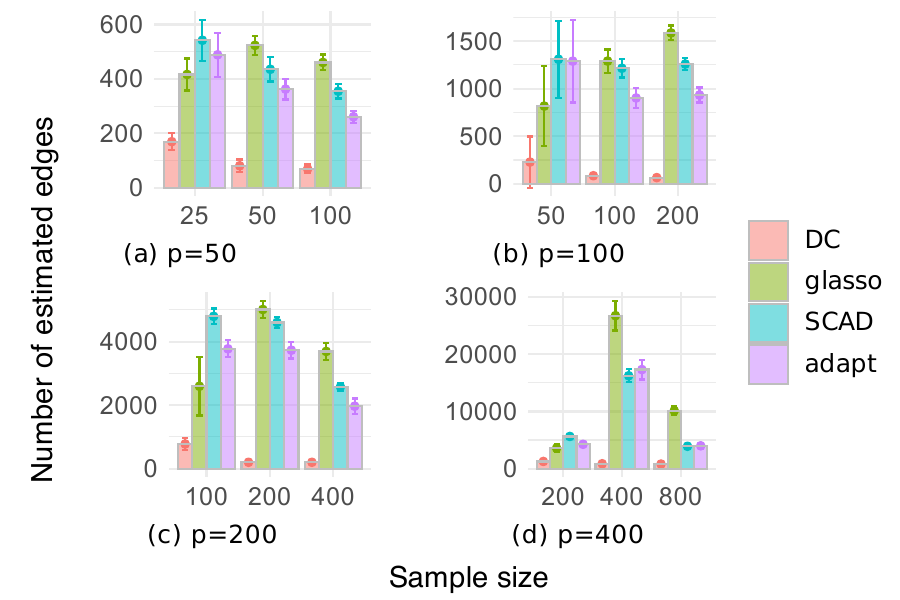}
\caption{\bf Number of edges selected through cross-validation for chain data}
\label{Fig:exp1_chain_numedge}
\end{figure}

Figures~\ref{Fig:exp1_chain} and~\ref{Fig:exp1_chain_numedge} show the F1 scores and the number of estimated non-zero edges for chain data, showing that our method performs well. Fig~\ref{Fig:exp1_chain} indicates that existing methods estimate denser graphs compared to random data, resulting in lower F1 scores. The estimated number of edges reveals that, compared to DC, conventional methods generally produce denser graphs. In contrast, DC estimates a relatively small and stable number of edges, consistent with the trends observed in random data.

\begin{figure}[ht]
\center
\includegraphics[width=\linewidth]{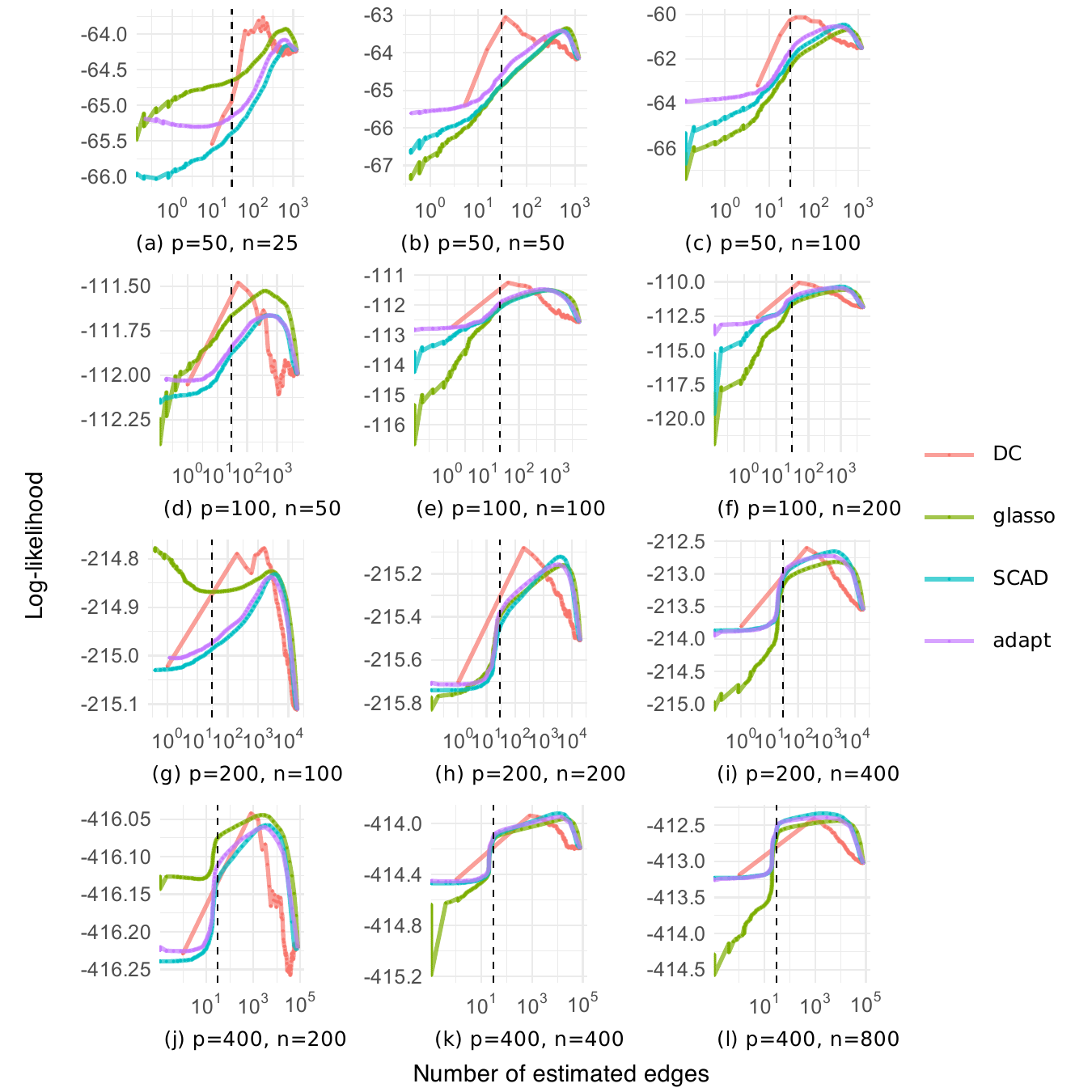}
\caption{\bf Number of selected edges and log-likelihood for chain data (black dotted line: true number of edges)}
\label{Fig:exp6_chain}
\end{figure}

Fig~\ref{Fig:exp6_chain} shows the relationship between the average number of selected edges and the average log-likelihood for chain data in cross-validation. DC often maximizes the average log-likelihood close to the true number of edges. However, similar to random data, the relationship with log-likelihood is not smooth, and there is a bias in the number of selected edges.

These results confirm that in experiments using cross-validation, our method selects a number of edges closer to the true number compared to existing methods. On the other hand, existing methods often select an extreme number of edges, resulting in lower F1 scores.

\subsection*{Experiment 2: Estimating a Fixed Number of Edges}
\begin{figure}[ht]
\centering
\includegraphics[width=\linewidth]{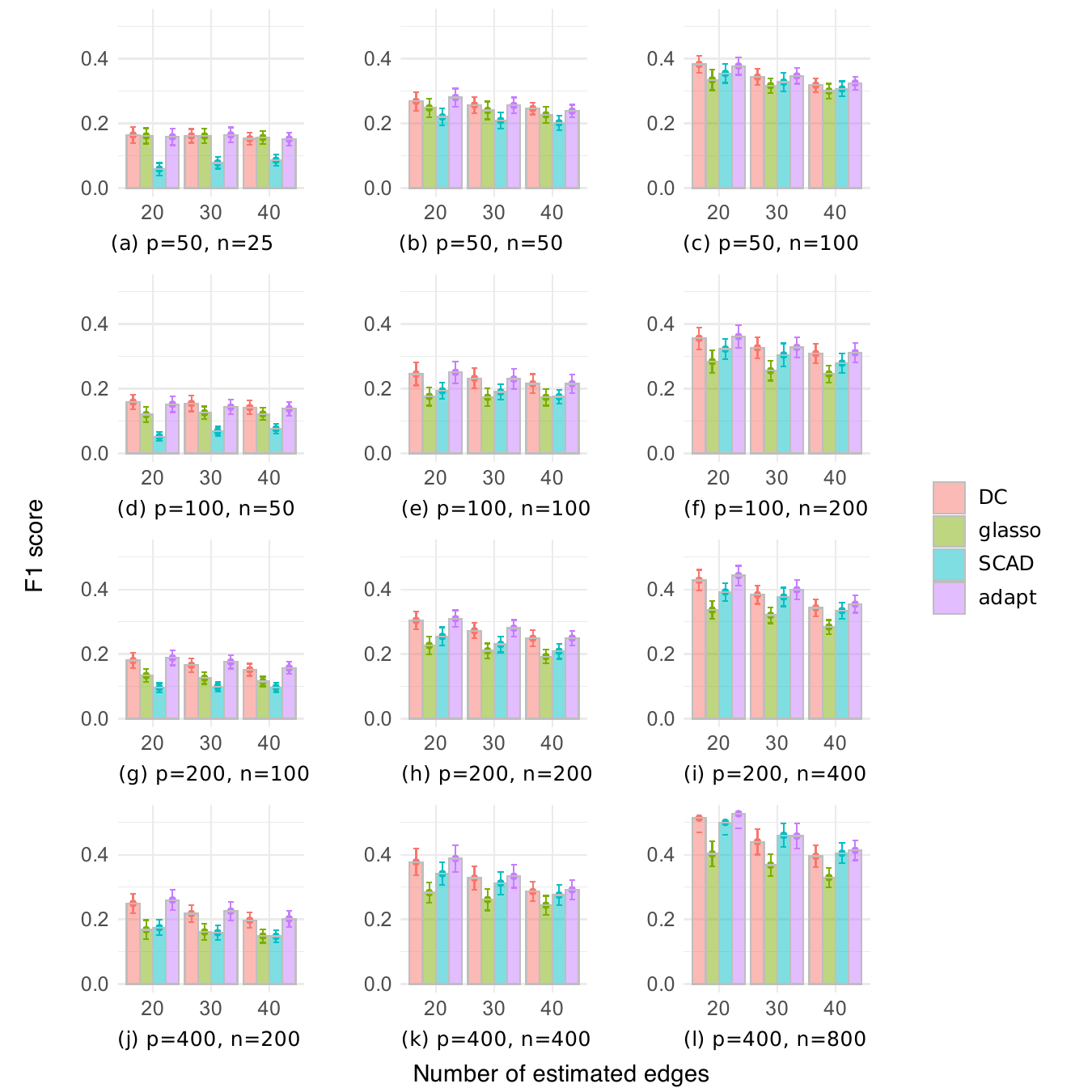}
\caption{\bf Average F1 score and $\pm2\sigma$ for each method on random data}
\label{Fig:exp2_random}
\end{figure}

Fig~\ref{Fig:exp2_random} shows the F1 scores for different numbers of estimated edges for random data. Overall, the score is better than in Experiment 1, with DC and adaptive lasso performing particularly well. Conversely, glasso and SCAD generally show lower score, especially when $p$ and $n$ are small. As the sample size increases, the score of all methods improves, possibly due to more precise estimation of the sample covariance matrix. Additionally, as the number of estimated edges increases, the score of all methods tends to decrease, which can be attributed to the increase in false positives.

\begin{figure}[ht]
\centering
\includegraphics[width=\linewidth]{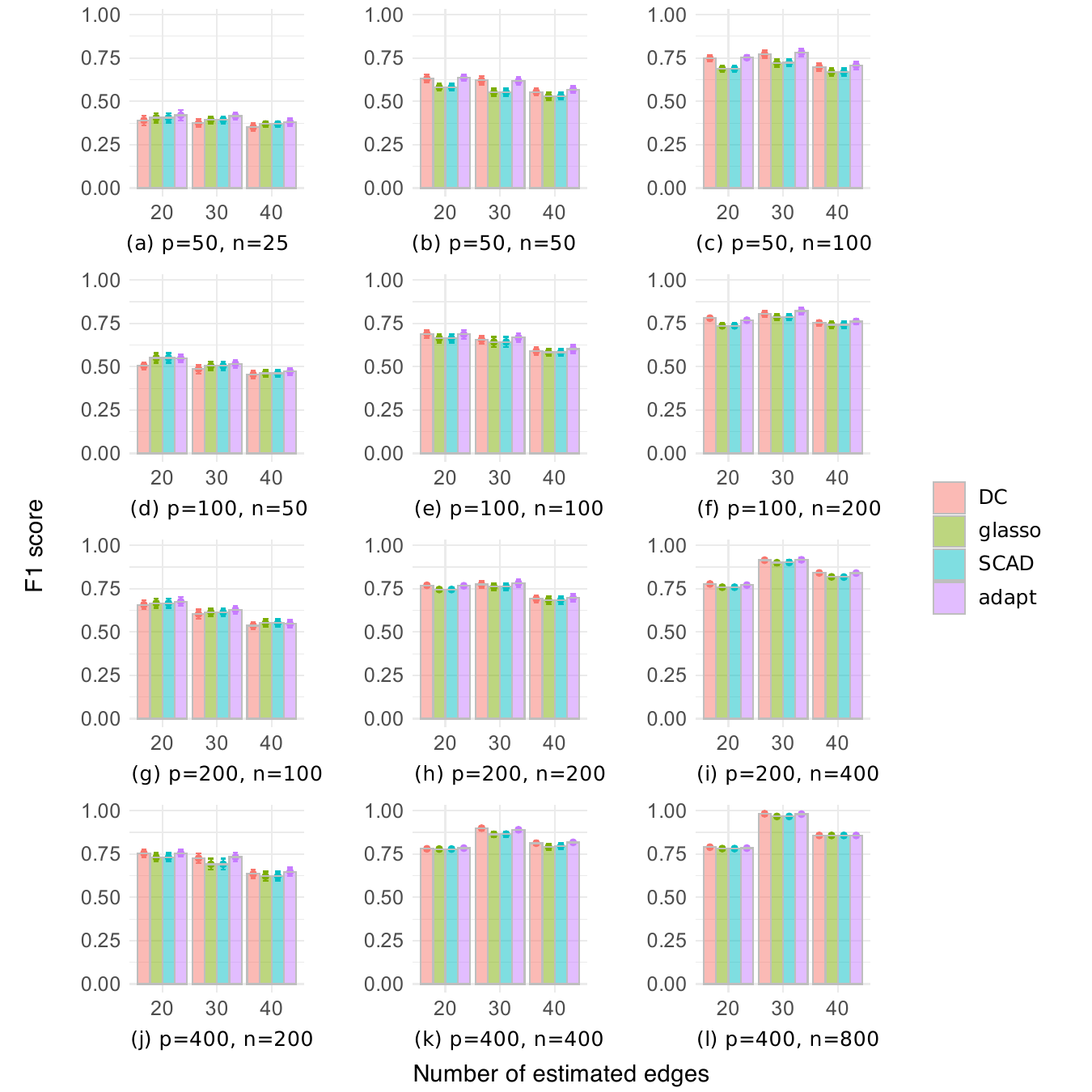}
\caption{\bf Average F1 score and $\pm2\sigma$ for each method on chain data}
\label{Fig:exp2_chain}
\end{figure}

Fig~\ref{Fig:exp2_chain} shows the F1 scores for different numbers of estimated edges for chain data. The scores are generally higher compared to random data, with our method and adaptive lasso showing slight superiority. For DC, score is comparable to or lower than other methods when $p < n$, but it performs comparably or better in other settings. Similar to the results for random data, score tends to decrease with an increase in the number of estimated edges for $p < n$. In other settings, the best results are often obtained when estimating the true number of 30 edges. These results suggest that chain data is easier to handle than random data, and estimating the true number of 30 edges leads to lower false positives and false negatives.

From these results, we confirmed that when estimating a fixed number of edges, our method and adaptive lasso perform better than other methods for random data. On the other hand, for chain data, all methods show very high scores with no significant differences.

\subsection*{Execution Time}
\begin{figure}[ht]
\centering
\includegraphics[width=\linewidth]{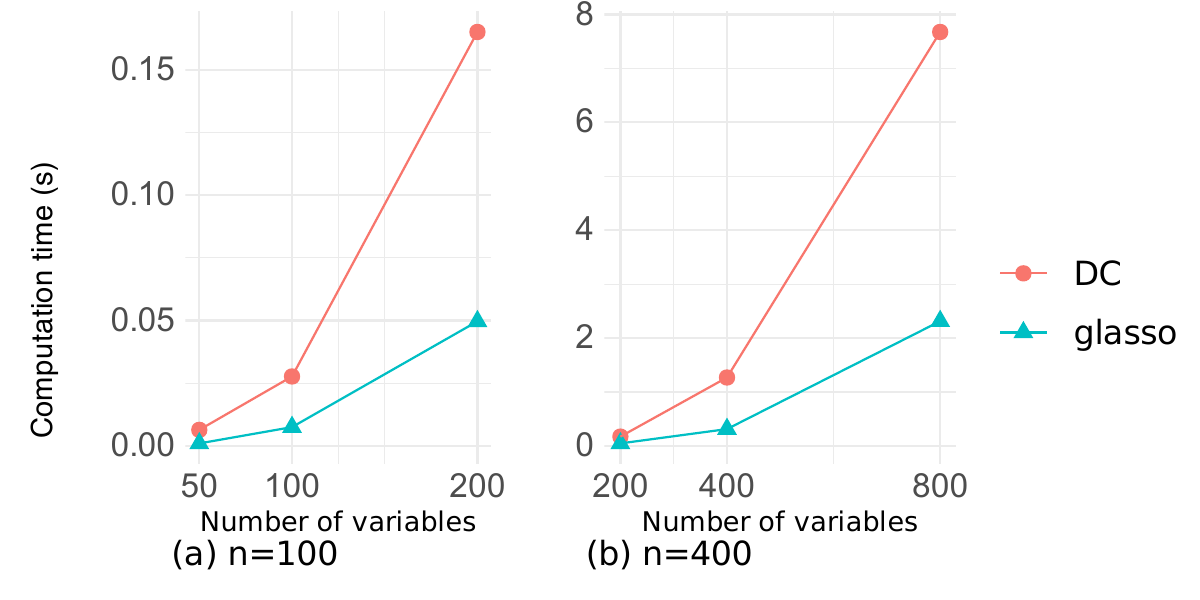}
\caption{\bf Relationship between the number of variables and execution time for random data}
\label{Fig:time_comparison_p_random}
\end{figure}

\begin{figure}[ht]
\center
\includegraphics[width=\linewidth]{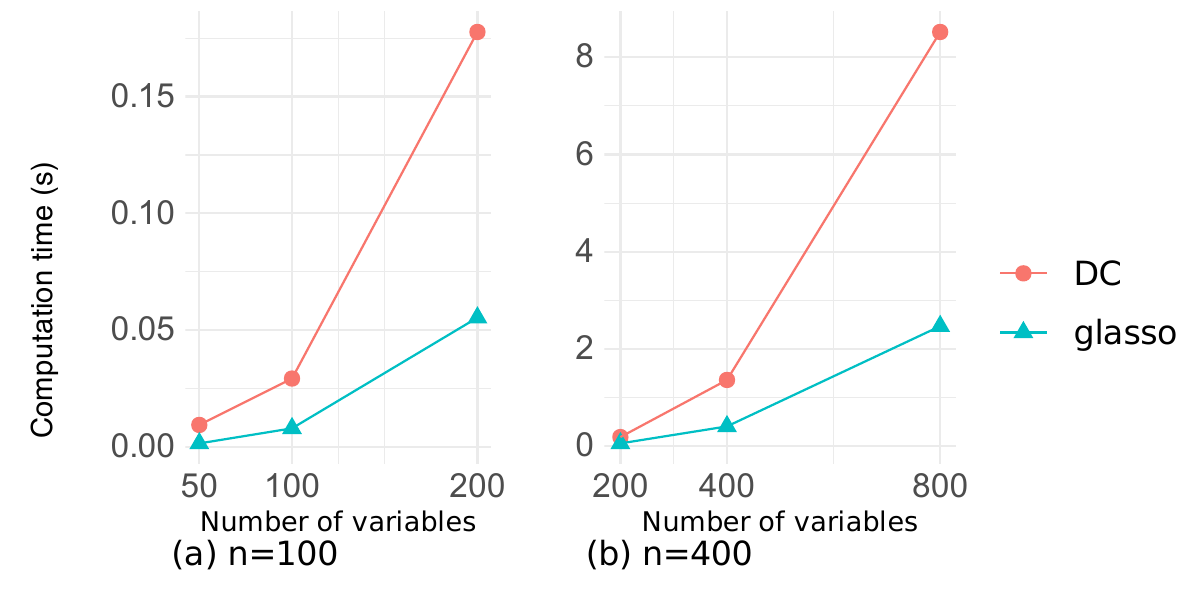}
\caption{\bf Relationship between the number of variables and execution time for chain data}
\label{Fig:time_comparison_p_chain}
\end{figure}

As shown in Algorithm~\ref{alg:dcglasso_alg}, our method involves several iterative processes, including the execution of glasso, making it more time-consuming than existing methods. Figures~\ref{Fig:time_comparison_p_chain} and~\ref{Fig:time_comparison_p_random} illustrate the relationship between the number of variables and the model estimation time for random and chain data, respectively, with (a) representing a sample size of 100 and (b) representing a sample size of 400. The data used includes artificial random and chain data with 30 non-zero components, and the average time over 10 runs with different initial random seeds was measured. For each model's hyperparameters, $K$ in our method was set to half the total number of edges ($K \leftarrow \frac{p(p-1)}{2} \times \frac{1}{2}$), and $\lambda$ in glasso was set to the median absolute value of the off-diagonal elements of the sample covariance matrix. Since there was minimal difference between existing methods, only the results of glasso are shown. Figures~\ref{Fig:time_comparison_p_chain} and~\ref{Fig:time_comparison_p_random} indicate that there is little difference in results between the data sets, and in all settings, our method takes about four times longer than glasso. This is due to the two reasons: 1. repeated execution of glasso in Algorithm \ref{alg:dcglasso_alg}, and 2. repeated eigenvalue calculations in the process of determining the penalty parameter $\eta$. However, for $p \leq 400$, both methods take less than 1.5 seconds to execute. Even for $p=800$, our method completes in approximately 8 seconds, demonstrating that the solution is computed sufficiently quickly.

\begin{figure}[ht]
\center
\includegraphics[width=\linewidth]{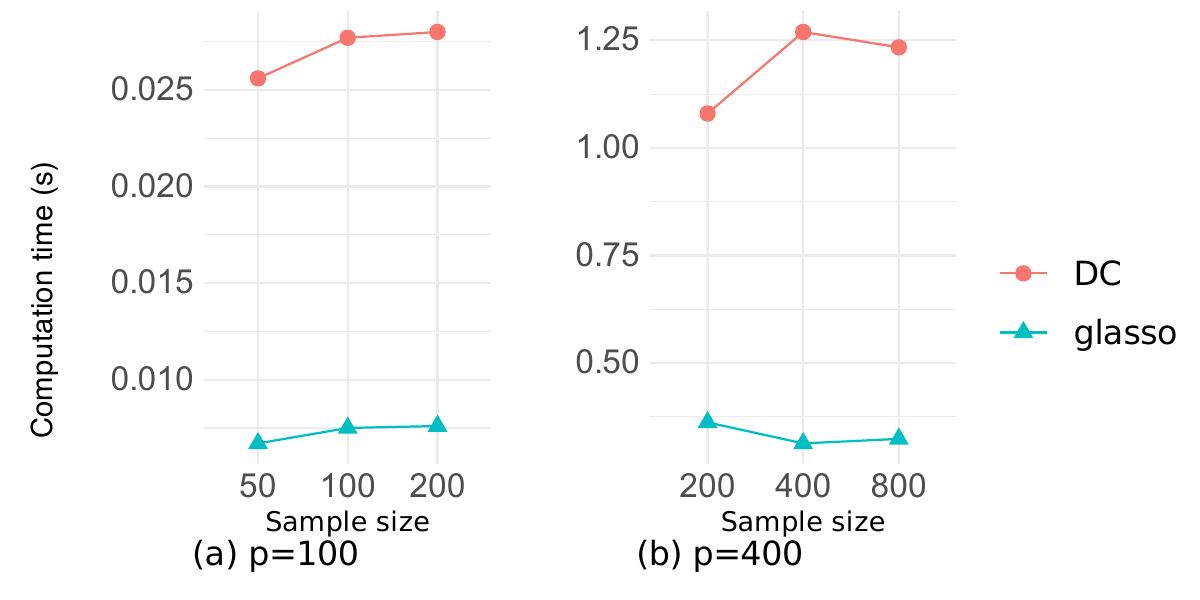}
\caption{\bf Relationship between sample size and execution time for random data}
\label{Fig:time_comparison_n_random}
\end{figure}

\begin{figure}[ht]
\center
\includegraphics[width=\linewidth]{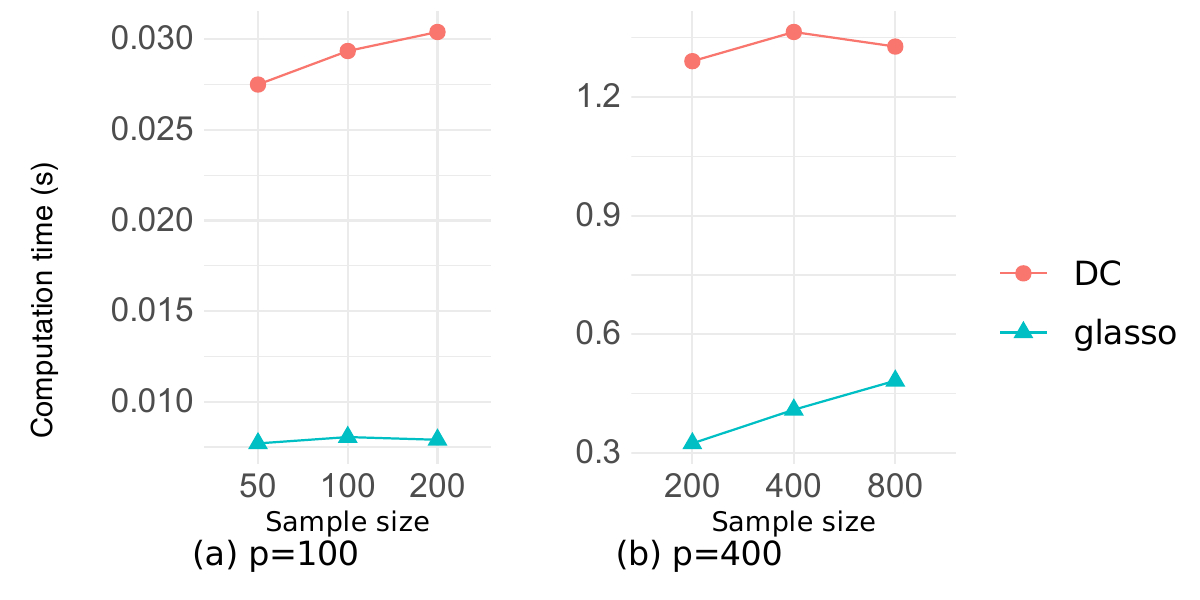}
\caption{\bf Relationship between sample size and execution time for chain data}
\label{Fig:time_comparison_n_chain}
\end{figure}

Figures~\ref{Fig:time_comparison_n_chain} and~\ref{Fig:time_comparison_n_random} show the same data as Figures~\ref{Fig:time_comparison_p_chain} and~\ref{Fig:time_comparison_p_random}, but with the horizontal axis representing sample size, with (a) showing the case for 100 variables and (b) for 400 variables. These figures confirm that the computation time for both methods is strongly dependent on the number of variables and hardly changes even if the sample size is multiplied several times.

\section*{Conclusion}
In this study, we proposed a sparse estimation method for Gaussian graphical models using the $\ell_0$ norm. We reformulated the problem with an $\ell_0$ norm constraint into a penalized form using the largest-$K$ norm and designed a DC algorithm to solve this problem.

To verify the performance of our method, we conducted two types of numerical experiments using two types of synthetic data. The first experiment, which involved estimating edges using cross-validation, confirmed that our method could estimate the true graph more accurately than existing methods. Examining the relationship between the average number of selected edges and the average log-likelihood in cross-validation, we found that our method often maximizes the likelihood near the true number of edges compared to existing methods. In the experiment where each model was trained to estimate a fixed number of edges, our method outperformed glasso and SCAD in score but was comparable to or slightly inferior to adaptive lasso. These results suggest that our method particularly excels in selecting true edges.

Regarding execution time, our method took approximately four times longer than glasso. This is due to two reasons: (1) our method repeatedly executes glasso, and (2) it repeatedly performs eigenvalue calculations in the process of determining the penalty parameter. However, the training time of our method was less than 1.5 seconds for $p \leq 400$ and about 8 seconds for $p = 800$, indicating that it can be solved quickly enough for practical use.

Future directions for this study include solving the problem formulated with cardinality constraints using other methods. For example, recent reports suggest that proximal gradient methods outperform DC algorithms in some aspects for discontinuous and non-convex sparse regression problems \cite{nakayama2021superiority}. Furthermore, since our method ultimately solves a penalized form of the problem, there may be cases where the solution does not satisfy the cardinality constraint. To strictly satisfy the cardinality constraint, a theoretical formulation of $K$ and $\eta$ based on the optimality conditions of the DC algorithm is necessary. To speed up the execution time of our method, it will be important to determine the penalty parameter $\eta$ more efficiently in Algorithm~\ref{alg:dcglasso_alg} and to use an accelerated version of glasso.


\bibliography{references}

\end{document}